\documentclass{article}%
\usepackage{amsmath}
\usepackage{amsfonts}
\usepackage{amssymb,float}
\usepackage{latexsym}
\usepackage[latin1]{inputenc}
\usepackage{xcolor}   
\usepackage[authoryear]{natbib}
\usepackage{amsmath}
\usepackage{amssymb,float}
\usepackage{latexsym}
\usepackage{epstopdf}
\usepackage{geometry}
\usepackage[active]{srcltx}
\usepackage{graphicx}
\usepackage{epstopdf}
\usepackage{enumerate}
\usepackage[ bookmarks=true,         bookmarksnumbered=true, colorlinks=true,
pdfstartview=FitV, linkcolor=blue, citecolor=blue, urlcolor=blue]{hyperref}
\usepackage{amssymb}%
\hypersetup{
    colorlinks = true,
    linkcolor = blue,
    citecolor = blue,
    urlcolor = blue
}
\providecommand{\U}[1]{\protect\rule{.1in}{.1in}}
\newtheorem{theorem}{Theorem}[section]

\newtheorem{corollary}{Corollary}[section]

\newtheorem{lemma}{Lemma}[section]

\newtheorem{remark}{Remark}[section]

\newenvironment{proof}[0]{\paragraph{Proof.}}{\rule{0.5em}{0.5em}}
\renewcommand\cite[1]{\citeauthor{#1} \citeyear{#1}}

\begin{document}
\title{Error analysis of generative adversarial network}

\maketitle

\begin{center}
\bigskip Mahmud Hasan, Hailin Sang

Department of Mathematics, University of Mississippi, University, MS, 38677,  USA\\
mhasan4@olemiss.edu, sang@olemiss.edu

 \bigskip
\end{center}
	
\begin{abstract}
The generative adversarial network (GAN) is an important model developed for high-dimensional distribution learning in recent years. However, there is a pressing need for a comprehensive method to understand its error convergence rate. In this research, we focus on studying the error convergence rate of the GAN model that is based on a class of functions encompassing the discriminator and generator neural networks. These functions are VC type with bounded envelope function under our assumptions, enabling the application of the Talagrand inequality. By employing the Talagrand inequality and Borel-Cantelli lemma, we establish a tight convergence rate for the error of GAN. This method can also be applied on existing error estimations of GAN and yields improved convergence rates. In particular, the error defined with the neural network distance is a special case error in our definition.
\end{abstract}

 \textbf{Keywords:} discriminator, error bound, GAN,  generator, Talagrand inequality, VC dimension.
 
 \section{Introduction}
  Generative adversarial networks (GANs) (Goodfellow et al. 2014; Li, Swersky and Zemel 2015; Dziugaite, Roy and Ghahramani 2015; Arjovsky, Chintala and Bottou 2017) have attracted much attention in machine learning and artificial intelligence communities in the past few years. As a powerful unsupervised method for learning and sampling from complex data distributions, GANs have revolutionized the field of deep learning by allowing the generation of highly realistic synthetic data and have achieved remarkable success in many machine learning tasks such as image synthesis, medical imaging and natural language generation (Radford, Metz and Chintala 2016; Reed et al. 2016; Zhu et al. 2017; Karras et al. 2018; Yi, Walia, and Babyn 2019). However, theoretical explanations for their empirical success need to be better established. Many problems on the theory and training dynamics of GANs are largely unsolved (Nowozin, Cseke, and Tomioka 2016; Arora et al. 2017; Liang 2021).

 One important issue in machine learning study is the generalization ability of neural network. Despite some impressive results, the generalization of GANs remains a challenging issue. The generalization of GANs refers to their ability to produce synthetic data similar to unseen actual data. In other words, it measures the quality of the generated data compared to the actual data distribution. This is critical, as GANs are increasingly being used in real-world applications, and their performance is essential to the success of these applications. This research aims to examine the generalization error of GANs and develops techniques to improve their performance. 

We consider the GAN models with both the generator and discriminator function classes parameterized. Let
$\mathcal{F}=\{f_{w}:\mathbb{R}^{p_0}\rightarrow\mathbb{R}\}$ be the discriminator function class realized by a neural network with parameter $w\in \mathcal{W}$ describing the weights of the network and $\mathcal{G}=\{g_{\theta}(z):\mathbb{R}^p\rightarrow\mathbb{R}^{p_0}\}$ be the generator neural network transformation class with weight parameter $\theta\in \Theta$.
Let the distribution of the random input variable $Z$ be $Z\sim\mu$ and the target distribution of $X$ be $X\sim\nu$. Denote $g_\theta^\mu$ as the probability distribution of $g_{\theta}(Z)$.
We write the  following objective function in the GAN model which is introduced in Goodfellow et al. (2014) and Arora et al. (2017),
\begin{equation}\label{sup1}
d_{\mathcal{F,\phi}}(g_\theta^\mu,\nu):=\sup _{w\in \mathcal{W} } \left\vert\mathbb{E}\phi (1-f_{w}(g_{\theta}(Z)))+\mathbb{E}\phi(f_{w}(X))\right\vert-2\phi(1/2),
\end{equation}
where $\phi$ is called the measuring function. We will require $\phi$ to be monotone increasing. Indeed, in practice training researchers often use $\phi(x)=x$, $\phi(x)=\log x$ or $\phi(x)=\log(\delta+(1-\delta)x)$ for some $0<\delta<1$. 
The objective is to minimize the above function and have  
\begin{equation}\label{inf1}\notag
\inf_{\theta\in \Theta} d_{\mathcal{F,\phi}}(g_\theta^\mu,\nu).
\end{equation}
 Suppose that we have $n$ independent and identical observations  $X_i\sim\nu, 1\le i\le n$,  and the generator produces $m$ independent and identical terms $g_{\theta}(Z_j)\sim g_\theta^\mu, 1\le j\le m$. We estimate the expectations by the empirical averages and minimize the following empirical version of \eqref{sup1},
\begin{equation*}
d_{\mathcal{F,\phi}}(g_\theta^{\hat{\mu}_m},\hat\nu_{n})=\sup_{w\in\mathcal{W}}\left\vert\hat{\mathbb{E}}_m\phi (1-f_w(g_{\theta}(Z)))+\hat{\mathbb{E}}_n \phi(f_w(X))\right\vert-2\phi(1/2),
\end{equation*}
i.e., 
 \begin{equation}\label{sup2}
d_{\mathcal{F,\phi}}(g_\theta^{\hat{\mu}_m},\hat\nu_{n})= \sup_{w\in\mathcal{W}}\left\vert\frac{1}{m}\sum_{j=1}^{m}\phi (1-f_{w}(g_{\theta}(Z_{j})))+\frac{1}{n}\sum_{i=1}^{n}\phi (f_{w}(X_{i}))\right\vert-2\phi(1/2).
\end{equation}
 Here 
$\hat\mu_{m}=\frac{1}{m}\sum_{j=1}^{m} \delta_{Z_j}$ and $\hat\nu_{n}=\frac{1}{n}\sum_{i=1}^{n}\delta_{X_i}$ are the corresponding empirical distributions of $\mu$ and $\nu$, respectively.
Equation \eqref{sup2} measures the distance between the empirical generator distribution $g_\theta^{\hat{\mu}_m}$ and the empirical target distribution $\hat\nu_{n}$. 
 
 In practice we train the above empirical objective function by minimizing $d_{\mathcal{F,\phi}}(g_\theta^{\hat{\mu}_m},\hat\nu_{n})$ over all $\theta\in \Theta$. However as the above GAN training is conducted it is important to understand the generalization error of GAN and obtain the convergence rates in a general setting. Here by generalization error we mean the difference between $\inf_{\theta\in\Theta}d_{\mathcal{F,\phi}}(g_\theta^{\hat{\mu}_m},\hat\nu_{n})$ and $\inf_{\theta\in \Theta} d_{\mathcal{F,\phi}}(g_\theta^\mu,\nu)$, i.e., the difference between the best empirical value of $d_{\mathcal{F,\phi}}(g_\theta^{\hat{\mu}_m},\hat\nu_{n})$ and the best population value of $d_{\mathcal{F,\phi}}(g_\theta^\mu,\nu)$. In the literature, the error analysis of the GAN model was studied with some different empirical objective functions, for example, the empirical objective function $d_{\mathcal{F,\phi}}(g^\mu_{\theta},\hat\nu_{n})$, 
\begin{equation}\label{sup3}
d_{\mathcal{F,\phi}}(g^\mu_{\theta},\hat\nu_{n})= \sup_{w\in\mathcal{W} }\left\vert\mathbb{E}\phi (1-f_{w}(g_{\theta}(Z)))+\frac{1}{n}\sum_{i=1}^{n}\phi (f_{w}(X_{i}))\right\vert-2\phi(1/2).
\end{equation}
There are a few research on error analysis of GAN model and most of the research is for the particular case $\phi(x)=x$ in which the objective function becomes the so called neural network distance. For example, Liang (2017) defined the error as the difference between the best value of \eqref{sup3} and the best value of \eqref{sup1}, i.e.,  $\inf_{\theta\in \Theta}d_{\mathcal{F,\phi}}(g^\mu_{\theta},\hat\nu_{n})-\inf_{\theta\in \Theta}d_{\mathcal{F,\phi}}(g_\theta^\mu,\nu)$, in the case of $\phi(x)=x$ for the objective function. We also notice that Liang (2017) only considered the  distribution $\nu$ of $X$ in Sobolev space. Similarly, in the case of $\phi(x)=x$ for the objective function, the error analysis in Zhang, Liu, Zhou, Xu, and He (2018) only  incorporates discriminator part and does not contain the role of generator. Ji, Zhou, and Liang (2021) captured the impact of both discriminator and generator neural networks. However the error in that paper is bounded by Rademacher complexity like the analysis in Zhang et al. (2018). Arora et al. (2017) made use of the $\epsilon$-net argument and showed error bound by taking difference between \eqref{sup2} and \eqref{sup1}.
 
Despite some advancements in the theoretical error analysis of GANs, significant limitations still persist. This research addresses these issues and introduces several key improvements for error analysis in GANs. Firstly, we propose a novel error definition for the general objective function studied in the literature (Arora et al. 2017). This definition encompasses two classes of functions, incorporating both the discriminator and generator neural network structures with bounded input variables and weight parameters, utilizing corresponding Lipschitz activation functions. Secondly, after finding the VC dimension and envelop function, we apply Talagrand inequality and Borel-Cantelli lemma to achieve tight convergence rates. Thirdly, we extend our error analysis technique to address similar cases in existing literature, obtaining tight convergence rates in those scenarios as well. We specifically apply our approach to errors studied in papers such as Arora et al. (2017), Liang (2017),  Zhang et al. (2018), and Ji et al. (2021), and thoroughly analyze the results. Furthermore, this research establishes the error for general objective functions, whereas in previous literature, the error was only defined for the specific case when $\phi(x)=x$. This broader scope allows for a more comprehensive and accurate analysis of GANs' error rates.

In the literature, researchers incorporate regularizations on neural networks to the objective functions, aiming to reduce the error, for example,   Oberman and Calder (2018) and Wei and Ma (2019).  In contrast, our paper focuses on an objective function without regularization. It is not clear whether such regularization techniques also contribute to minimizing the error in GAN models. We leave the investigation of the regularized case for future research. 

The paper is organized as follows. In Section \ref{mains}, we introduce the discriminator and generator neural network structure in the GAN model and establish the main results.  
The proofs of the main results are given in Section \ref{proof}. 
We extend our technique to other type error bounds and compare our results with existing ones in Section \ref{exten}. 
We conclude and discuss future work in Section \ref{conc}.

\section{ Main results}\label{mains}
Given a vetor $w\in \mathbb{R}^p$, $\lVert w\rVert:=\lVert w\rVert_2:=\left(\sum_{i=1}^{p}w_{i}^2\right)^{1/2}$ denotes the $l_{2}$ norm and $\lVert w\rVert_1=\sum_{i=1}^{p}\lvert w_{i}\rvert$ denotes the $l_{1}$ norm, where  $w_{i}$ is the $i$-th coordinate of $w$.
We use $\lVert W\rVert:=\lVert W\rVert_{F}:=\left( \sum_{ij}W_{ij}^2\right)^{1/2}$ as the Frobenius norm of the matrix $W=[W_{ij}]$.
For the discriminator and generator neural networks, we consider the following compact parameter spaces 
 \begin{align}\label{W}
\mathcal{W}=\prod_{i=1}^{d}\{W_{i}\in{\mathbb{R}}^{p_{i}\times p_{i-1}}: p_d=1, \lVert W_{i}\rVert\leq M_{w}(i),\;\; \text{for some}\;\; M_{w}(i)>0\},\notag\\			
\Theta=\prod_{j=1}^{s}\{\Theta_{j}\in {\mathbb{R}}^{q_{j}\times q_{j-1}}: q_0=p, q_s=p_0, \lVert \Theta_{j}\rVert\leq M_{\theta}(j),\;\; \text{for some}\;\; M_{\theta}(j)>0\}.
\end{align}

The discriminator  class of functions $ \mathcal{F}:= \{ f_{w}(x): w\in \mathcal{W} \}$ is taken as
\begin{equation}\label{F}
\mathcal{F}=\{f_{w}(x): f_{w}(x)= W_{d}\sigma_{d-1}(W_{d-1}\sigma_{d-2}(\cdots\sigma_{1}(W_{1}x))), w\in \mathcal{W}, x\in \mathbb{R}^{p_0}\},  
\end{equation}
where $d$ is the number of layers of the network and the number of weights is $\sum_{i=1}^{d}\lVert W_{i}\rVert_{0}$, $\lVert W_{i}\rVert_{0}$ denotes the number of non-zero entries of a matrix or vector.
$\sigma_{i}$ denotes activation function for $i=1, 2, \cdots, d-1$ and it works on a vector term wise. We assume each $\sigma_{i}$ is $L_{w}(i)$-Lipschitz, i.e., 
\begin{align}\label{sigmaL}
\lVert \sigma_{i}(r_{1})-\sigma_{i}(r_{2}) \rVert \leq L_{w}(i)\lVert r_{1}-r_{2} \rVert
\end{align}
for some $L_{w}(i)>0$ and for any $ r_{1}, r_{2} \in \mathbb{R}^{p_i}$.

The generator class of functions $ \mathcal{G}:= \{ g_{\theta}(z):\theta\in \Theta \}$ is taken as the form 
\begin{equation}\label{G}
\mathcal{G}=\{g_{\theta}(z):  g_{\theta}(z)= \Theta_{s}\phi_{s-1}(\Theta  _{s-1}\phi_{s-2}(\cdots\phi_{1}(\Theta_{1}z))), \theta\in \Theta, z\in \mathbb{R}^p \}, 
\end{equation}
where $s$ is the number of layers of the network, the number of weights is $\sum_{j=1}^{s}\lVert\Theta_{j}\rVert_{0}$, $\lVert\Theta_{j}\rVert_{0}$ denotes the number of non-zero entries of a matrix or vector and $\phi_{j}$ denotes activation function for $j=1, 2, \cdots, s-1$ and it works on a vector term wise. We assume each $\phi_{j}$ is $L_{\theta}(j)$-Lipschitz, i.e., 
\begin{align}\label{phiL}
\lVert \phi_{j}(r_{1})-\phi_{j}(r_{2}) \rVert \leq L_{\theta}(j)\lVert r_{1}-r_{2} \rVert
\end{align}
 for some $L_{\theta}(j)>0$ and for any $ r_{1},r_{2} \in \mathbb{R}^{q_j}$.

We can write each function $f_{w}(g_{\theta}(z))$ 
as a concatenation of the discriminator and generator neural networks in the form as

 \begin{equation}\label{FG}
f_{w}(g_{\theta}(z)) =W_{d}\sigma_{d-1}(W_{d-1}\sigma_{d-2}(\cdots\sigma_{1}(W_{1} \Theta_{s}\phi_{s-1}(\Theta  _{s-1}\phi_{s-2}(\cdots\phi_{1}(\Theta_{1}z)))))) 
\end{equation}
 with the number of layers $d+s-1$ and the number of weights $\sum_{i=1}^{d}\lVert W_{i}\rVert_{0}+\sum_{j=1}^{s}\lVert\Theta_{j}\rVert_{0}$. See Petersen (2022).
  Denote the class of functions of $f_{w}(g_{\theta}(z))$ as $ \mathcal{F}\circ \mathcal{G}$.  

Finally, let the new class of functions indexed by $ u:=(\theta,w)\in U:=(\Theta, \mathcal{W})$ and  $w$ be
\begin{align}\label{H}
\mathcal{H}=\{h_{u}:h_{u}(z)= \phi(1-f_{w}(g_{\theta}(z)))\}
\end{align}
and 
\begin{align}\label{F1}
\mathcal{F}_{1}=\{f_{1}:f_{1}(x)=\phi(f_{w}(x))\}.
\end{align}
We can write $\mathcal{H}=\phi(1-\mathcal{F}\circ \mathcal{G})$ and  $\mathcal{F}_{1}=\phi(\mathcal{F})$.

Here is the main result of the paper. 
\begin{theorem}\label{theorem1}
Let $\mathcal{P}_{\chi}$ be the class of probability measures over the compact domain $\chi$,  $\mathcal{F}$ and $\mathcal{G}$ be the discriminator and generator classes of functions defined in (\ref{F}) and (\ref{G}) respectively. Assume that the measuring function $\phi$ is monotone increasing and the random input variable $Z$ satisfies $\|Z\|<\infty$.  Let the target distribution function be $\nu\in\mathcal{P}_{\chi}$. Then
\begin{equation*}
\left \vert \inf\limits_{\theta\in\Theta}d_{\mathcal{F,\phi}}(g_\theta^{\hat{\mu}_m},\hat\nu_{n})-\inf\limits_{\theta\in\Theta}d_{\mathcal{F,\phi}}(g_\theta^{{\mu}},\nu)\right\vert=O_{a.s}\left(\left(\frac{\log m}{m}\right)^\frac{1}{2}\right)+ O_{a.s}\left(\left(\frac{\log n}{n}\right)^\frac{1}{2}\right).
\end{equation*}	
\end{theorem}
\section{Proof}\label{proof}
First of all, we have the following lemma which switches the problem in Theorem \ref{theorem1} to an empirical process problem. 
\begin{lemma}\label{L2.1}
Under the same setup and assumptions as in Theorem \ref{theorem1}, we have 
\begin{equation}\label{Lem1}
\left \vert \inf\limits_{\theta\in\Theta}d_{\mathcal{F,\phi}}(g_\theta^{\hat{\mu}_m},\hat\nu_{n})-\inf\limits_{\theta\in\Theta}d_{\mathcal{F,\phi}}(g_\theta^{{\mu}},\nu)\right\vert\leq \sup_{u\in U} \left \vert\frac{1}{m}\sum_{j=1}^{m} \left( h_{u}(Z_{j})-\mathbb{E}h_{u}(Z_{j})\right)\right\vert 
+\sup_{w\in \mathcal{W}} \left \vert\frac{1}{n}\sum_{i=1}^{n} \left( f_{1}(X_{i})-\mathbb{E}f_{1}(X_{i})\right)\right\vert 
\end{equation}
where $h_{u}(Z)= \phi(1-f_{w}(g_{\theta}(Z)))$, $u=(\theta,w)$ and $f_{1}(X)=\phi(f_{w}(X))$.
\end{lemma}

\begin{proof}
We are bounding the error $\left \vert \inf\limits_{\theta\in\Theta}d_{\mathcal{F,\phi}}(g_\theta^{\hat{\mu}_m},\hat\nu_{n})-\inf\limits_{\theta\in\Theta}d_{\mathcal{F,\phi}}(g_\theta^{{\mu}},\nu)\right\vert$. The standard properties of supremum and infimum are used to get the bound. The discriminator and generator sample sizes are taken to be $n$ and $m$, and we derive new classes of functions which are the discriminator class and the class of compositions of the discriminator and generator.	
By applying $\lvert \inf f(\cdot)-\inf g(\cdot)\rvert \leq \sup \lvert f(\cdot)-g(\cdot)\rvert$ and $\sup \lvert f(\cdot)+g(\cdot)\rvert\le \lvert  \sup f(\cdot)+\sup g(\cdot)\rvert$, we have 
\begin{align*}
&\left \vert \inf\limits_{\theta\in\Theta}d_{\mathcal{F,\phi}}(g_\theta^{\hat{\mu}_m},\hat\nu_{n})-\inf\limits_{\theta\in\Theta}d_{\mathcal{F,\phi}}(g_\theta^{{\mu}},\nu)\right\vert\\
&=\left\vert \inf_{\theta\in \Theta} \left\{ \sup_{w\in\mathcal{W}}\left\vert\hat{\mathbb{E}}_m\phi (1-f_w(g_{\theta}(Z)))+\hat{\mathbb{E}}_n \phi(f_w(X))\right\vert-2\phi(1/2) \right\} \right.\\
&\left.-\inf_{\theta\in \Theta}\left\{ \sup_{w\in \mathcal{W}} \left\vert{\mathbb{E}}\phi (1-f_w(g_{\theta}(Z)))+{\mathbb{E}} \phi(f_w(X))\right\vert-2\phi(1/2)\right\}\right\vert \\
&\leq \sup_{\theta\in \Theta} \left\vert \sup_{w\in\mathcal{W}}\left\vert\frac{1}{m}\sum_{j=1}^{m}\phi (1-f_{w}(g_{\theta}(Z_{j})))+\frac{1}{n}\sum_{i=1}^{n}\phi (f_{w}(X_{i}))\right\vert - \sup_{w\in \mathcal{W}} \left\vert{\mathbb{E}}\phi (1-f_w(g_{\theta}(Z)))+{\mathbb{E}} \phi(f_w(X))\right\vert \right\vert \\
&\leq\sup_{\theta\in \Theta} \sup_{w\in \mathcal{W}}\left \vert \frac{1}{m}\sum_{j=1}^{m}\phi (1-f_{w}(g_{\theta}(Z_{j})))+\frac{1}{n}\sum_{i=1}^{n}\phi (f_{w}(X_{i}))- [{\mathbb{E}}\phi (1-f_w(g_{\theta}(Z)))+{\mathbb{E}} \phi(f_w(X))]\right\vert\\
&\leq \sup_{\theta\in \Theta} \sup_{w\in \mathcal{W}}\left \vert \frac{1}{m}\sum_{j=1}^{m}\phi(1-f_{w}(g_{\theta}(Z_{j})))- \mathbb{E}\phi (1-f_{w}(g_{\theta}(Z)))\right\vert+ \sup_{w\in \mathcal{W}}\left \vert \frac{1}{n}\sum_{i=1}^{n}\phi(f_{w}(X_{i}))-\mathbb{E}\phi(f_{w}(X))\right\vert\\
&=\sup_{u\in U} \left \vert\frac{1}{m}\sum_{j=1}^{m} \left( h_{u}(Z_{j})-\mathbb{E}h_{u}(Z_{j})\right)\right\vert 
+\sup_{w\in \mathcal{W}} \left \vert\frac{1}{n}\sum_{i=1}^{n} \left( f_{1}(X_{i})-\mathbb{E}f_{1}(X_{i})\right)\right\vert.
\end{align*} 
\end{proof}

The function $h_{u}(z)= \phi(1-f_{w}(g_{\theta}(z)))$ with index $u=(\theta,w)$ is related to the composition function $f_{w}(g_{\theta}(z))$ in which the generator is composed into the
discriminator in the formulation of the objective function in the
GAN training. The function $f_{1}=\phi(f_{w}(x))$ with index $w$ involves the  discriminator function $f_{w}(x)$.

\subsection{VC-dimensions of $\mathcal{H}$ and $\mathcal{F}_{1}$}

In this section, we are going to show that $\mathcal{H}$ and $\mathcal{F}_{1}$ are VC classes of functions and find their VC-dimensions. For this purpose, we need to find the VC-dimension of  $\mathcal{F}$ and the VC-dimension of $ \mathcal{F}\circ \mathcal{G}$.

For a class of functions generated from a neural network, the VC-dimension is affected by various attributes of the network: the number of non-zero parameters (i.e., weights and biases) $W$, 
and the number of layers $L$.
The best upper bounds are $O(W^2)$ (Goldberg and Jerrum 1995) if we only use $W$ as the parameter and $O(WL\log W+W L^{2}
)$ (Bartlett, Maiorov, and Meir 1998) if we use $W$ and $L$ as the parameters, both of which hold for piece-wise polynomial activation functions. Bartlett, Harvey, Liaw, and Mehrabian (2019) improved the VC-dimension upper bounds to $O(W L\log W) $  for the special case in which the activation functions are piece-wise linear functions including the ReLU activation function. In our case, the three classes of functions given in \eqref{F}, \eqref{G} and \eqref{FG} depend on the weight parameters and number of layers. Besides, we are applying VC-dimension upper bounds as  $O(W L\log W) $ which is lower than $O(W^2)$ and $O(WL\log W+W L^{2})$.

\begin{lemma}\label{lemma2}
Let $\mathcal{H}=\phi(1-\mathcal{F}\circ \mathcal{G})$ and $\mathcal{F}_{1}=\phi(\mathcal{F})$ be the classes of functions formulated by $\mathcal{F}\circ \mathcal{G}$ and $\mathcal{F}$ which were defined in equation \eqref{FG} and \eqref{F} and $\phi$ is the monotone function. Then 

\begin{equation}\label{VCH}\notag
v(\mathcal{H})=O\left(\left(d+s-1\right)\left(\sum_{i=1}^{d}\lVert W_{i}\rVert_{0}+\sum_{j=1}^{s}\lVert \Theta_{j}\rVert_{0}\right)\log\left(\sum_{i=1}^{d}\lVert W_{i}\rVert_{0}+\sum_{j=1}^{s}\lVert \Theta_{j}\rVert_{0}\right)\right),
\end{equation}
\begin{align}\label{VCF1}
v(\mathcal{F}_{1})=O\left(d\sum_{i=1}^{d}\lVert W_{i}\rVert_{0}\log\left(\sum_{i=1}^{d}\lVert W_{i}\rVert_{0}\right)\right),
\end{align}
where $v(\mathcal{H})$ and $v(\mathcal{F}_{1})$ are VC-dimensions of $\mathcal{H}$ and $\mathcal{F}_{1}$.
\end{lemma}

\begin{proof}
 	
 For the class of function $\mathcal{F}$, the numbers of layers and  weights are  $d$ and $\sum_{i=1}^{d}\lVert W_{i}\rVert_{0}$ respectively from equation \eqref{F}. For $\mathcal{F}\circ \mathcal{G}$ the numbers of layers and weights are
 $(d+s-1)$ and $\sum_{i=1}^{d}\lVert W_{i}\rVert_{0}+\sum_{j=1}^{s}\lVert\Theta_{j}\rVert_{0}$ respectively from equation \eqref{FG}.
 Besides, We apply the piecewise linear function including ReLU function as the activation functions, $\sigma_{i}$, $1\le i\le d-1$,  and $\phi_{j}$, $1\le j\le s-1$,  for  both the discriminator and generator neural networks. 
Then using  Bartlett et al. (2019) the VC-dimension for $\mathcal{F}$ is 
\begin{align}\label{VCF}\notag
v(\mathcal{F})=O\left(d\sum_{i=1}^{d}\lVert W_{i}\rVert_{0}\log\left(\sum_{i=1}^{d}\lVert W_{i}\rVert_{0}\right)\right).
\end{align}
Similarly the VC-dimension for $\mathcal{F}\circ \mathcal{G}$ is
\begin{equation}\label{VCFG}\notag
v(\mathcal{F}\circ \mathcal{G})=O\left(\left(d+s-1\right)\left(\sum_{i=1}^{d}\lVert W_{i}\rVert_{0}+\sum_{j=1}^{s}\lVert \Theta_{j}\rVert_{0}\right)\log\left(\sum_{i=1}^{d}\lVert W_{i}\rVert_{0}+\sum_{j=1}^{s}\lVert \Theta_{j}\rVert_{0}\right)\right).
\end{equation}
Since $\phi$ is monotone, so $\mathcal{F}_1$ is VC class with VC-dimension $ v(\mathcal{F}_1)=v(\mathcal{F})$ (see, e.g.,  Lemma 2.6.18 in Vaart and Wellner 1996).
Similarly the VC-dimension for $\mathcal{H}$ is
\begin{equation*}
v(\mathcal{H})=v(\mathcal{F}\circ \mathcal{G})=O\left(\left(d+s-1\right)\left(\sum_{i=1}^{d}\lVert W_{i}\rVert_{0}+\sum_{j=1}^{s}\lVert \Theta_{j}\rVert_{0}\right)\log\left(\sum_{i=1}^{d}\lVert W_{i}\rVert_{0}+\sum_{j=1}^{s}\lVert \Theta_{j}\rVert_{0}\right)\right).
\end{equation*}
\end{proof}

\subsection{Envelope functions  of $\mathcal{H}$ and $\mathcal{F}_{1}$}

In this section, we evaluate the envelope function for the class  
$\mathcal{H}$ and $\mathcal{F}_{1}$ by applying the boundedness of the weight parameter matrices given by equation \eqref{W}  and the Lipschitz property of the activation functions $\sigma_{i}$, $\phi_{j}$, $1\le i\le d-1$, $1\le j\le s-1$.
\begin{lemma}\label{lemma3}
Let $\mathcal{H}$ and $\mathcal{F}_{1}$ be the classes of functions which were defined in \eqref{H} and \eqref{F1}. Under the assumptions as in Theorem \ref{theorem1},  for any $h_{u}\in\mathcal{H}$ and $f_{1}\in\mathcal{F}_{1}$
\begin{align}\label{ENFP}
\lvert f_{1}(X)\rvert\leq \max\{\lvert\phi(K_{1})\rvert,\lvert\phi(-K_{1})\rvert\}:=K_{3},
\end{align}
\begin{align*}
\lvert h_{u}(Z)\rvert\leq \max\{\lvert\phi(1-K_{2})\rvert,\lvert\phi(1+K_{2})\rvert\}:=K_{4},
\end{align*}                                                           
where $K_{1}=\prod_{i=1}^{d} M_{w}(i) \prod_{i=1}^{d-1} L_{w}(i)B_X$ and $K_{2}= \prod_{i=1}^{d}M_{w}(i) \prod_{i=1}^{d-1} L_{w}(i)\prod_{j=1}^{s}M_{\theta}(j) \prod_{j=1}^{s-1} L_{\theta}(j)B_Z$, $\|X\|<B_X, \|Z\|<B_Z$. 
\end{lemma}
\begin{proof}
We apply the boundedness of $\mathcal{F}$ and $\mathcal{F}\circ \mathcal{G}$ to obtain the bound of $\mathcal{F}_{1}$ and $\mathcal{H}$. For any $f_{w}\in \mathcal{F}$, we have   
\begin{align}
\lvert f_{w}(X)\rvert= & \lvert W_{d}\sigma_{d-1}(W_{d-1}\sigma_{d-2}(....\sigma_{1}(W_{1}X)))  \rvert \notag\\
\leq &  M_{w}(d)\lVert \sigma_{d-1}(W_{d-1}\sigma_{d-2}(....\sigma_{1}(W_{1}X))) \rVert \label{CS}\\
\leq & M_{w}(d) L_{w}(d-1)\lVert (W_{d-1}\sigma_{d-2}(....\sigma_{1}(W_{1}X))) \rVert,   \label{fX}
\end{align}
where \eqref{CS} follows from Cauchy-Schwarz inequality and \eqref{fX} follows the fact that $\sigma_{d-1}(\cdot)$ is $L_{w}(d-1) $-Lipschitz \eqref{sigmaL}. Repeating the process we obtain 

\begin{equation}\label{ENF}
\lvert f_{w}(X)\rvert \leq \prod_{i=1}^{d} M_{w}(i) \prod_{i=1}^{d-1} L_{w}(i) B_X:=K_{1}.
\end{equation}

\begin{equation}
\lvert f_{1}(X)\rvert \leq \max\{\lvert\phi(K_{1})\rvert,\lvert\phi(-K_{1})\rvert\}= K_{3} \notag.
\end{equation}

Similarly 
\begin{align}
\lvert f_{w}(g_{\theta}(Z))\rvert= & \lvert W_{d}\sigma_{d-1}(W_{d-1}\sigma_{d-2}(....\sigma_{1}(W_{1} \Theta_{s}\phi_{s-1}(\Theta_{s-1}\phi_{s-2}(....\phi_{1}(\Theta_{1}Z))))))\rvert \notag\\
\leq &\prod_{i=1}^{d}M_{w}(i) \prod_{i=1}^{d-1} L_{w}(i) \lVert ( \Theta_{s}\phi_{s-1}(\Theta_{s-1}\phi_{s-2}(....\phi_{1}(\Theta_{1}Z))))\rVert \notag\\
\leq&\prod_{i=1}^{d}M_{w}(i) \prod_{i=1}^{d-1} L_{w}(i) M_{\theta}(s) \lVert (\phi_{s-1}(\Theta_{s-1}\phi_{s-2}(....\phi_{1}(\Theta_{1}Z))))\rVert \notag\\
\leq&\prod_{i=1}^{d}M_{w}(i) \prod_{i=1}^{d-1} L_{w}(i)M_{\theta}(s)L_{\theta}(s-1) \lVert (\Theta_{s-1}\phi_{s-2}(....\phi_{1}(\Theta_{1}Z))))\rVert, \label{fgx}
\end{align}

where \eqref{fgx} follows from the fact that $\phi_{s-1}(\cdot) $ is $L_{s-1} $-Lipschitz \eqref{phiL}. Repeating the process we obtain
\begin{equation}\label{ENFG}
 \lvert f_{w}(g_{\theta}(Z))\rvert\leq 
 \prod_{i=1}^{d}M_{w}(i) \prod_{i=1}^{d-1} L_{w}(i)\prod_{j=1}^{s}M_{\theta}(j) \prod_{j=1}^{s-1} L_{\theta}(j)B_Z:=K_{2}.
\end{equation}
 We can write $1-K_{2}\le 1-f_{w}(g_{\theta}(Z))\le 1+K_{2}$.\\
Since $\phi$ is monotone,
\begin{align}
 \lvert h_{u}(Z)\rvert=\lvert\phi(1-f_{w}(g_{\theta}(Z)))\rvert
 \le \max\{\lvert\phi(1-K_{2})\rvert,\lvert\phi(1+K_{2})\rvert\}= K_{4} \notag.
\end{align}
\end{proof}


\subsection{Talagrand inequality and convergence rates}
Lemma \ref{lemma2} and Lemma \ref{lemma3} show that the VC-dimension and the bound of $\mathcal{H}$ are given by  $v(\mathcal{H})$ and $K_{4}$  respectively. So for the collection of uniformly bounded measurable function $\mathcal{H}$ on  $(S,\mathcal{S})$ with $S=\mathbb{R}^p$, $\mathcal{H}$ is VC type and 
there exists a constant $A_{1}$,  for all probability measure $Q$ on $\mathcal{S}$, we have,

\begin{align}\label{N1H}
N_{1}(\mathcal{H},L_{2}(Q),\epsilon)\leq \left(\frac{A_{1}\lVert K_{4} \rVert_{L_{2}(Q)}}{\epsilon}\right)^{v(\mathcal{H})}, \;0<\epsilon< 1,
\end{align}
where $\lvert h_u(Z)\rvert\leq K_{4}$ for all $ h\in \mathcal{H}$ and $\mathcal{S}$ is the Borel $\sigma$-algebra on $S$.\\
Similarly, the VC-dimension and the bound of $\mathcal{F}_{1}$ are $v(\mathcal{F}_{1})$ and $K_{3}$. For the collection of uniformly bounded measurable functions on $(S_0,\mathcal{S}_0)$ with $S_0=\mathbb{R}^{p_{0}}$ and $\mathcal{S}_0$ is the Borel $\sigma$-algebra on $S_0$, $\mathcal{F}_{1}$ is VC type and 
there exists a constant $A_{2}$ for all probability measure $Q_0$ on $\mathcal{S}_0$, we have, 
\begin{align}\label{N2F1}
N_{2}(\mathcal{F}_{1},L_{2}(Q_0),\epsilon)\leq \left(\frac{A_{2}\lVert K_{3} \rVert_{L_{2}(Q_0)}}{\epsilon}\right)^{v(\mathcal{F}_{1})}, \;0<\epsilon< 1 .
\end{align}

In \eqref{N1H} and \eqref{N2F1}, $N_{1}(\mathcal{H},L_{2}(Q),\epsilon)$ and $N_{2}(\mathcal{F}_{1},L_{2}(Q_0),\epsilon)$ denote the smallest number of $L_{2}(Q)$-balls or $L_{2}(Q_0)$-balls of radius at most $\epsilon$ required to cover $\mathcal{H}$ and $\mathcal{F}_{1}$, see, e.g.,  Pe\~{n}a and Gin\'{e} (1999).

Talagrand inequality (Talagrand 1996) is an important exponential inequality to provide a uniform convergence rate for an empirical process. We have the following version of Talagrand inequality (see, e.g., Einmahl and Mason 2000; Gin\'{e} and Guillou 2001, 2002; Gin\'{e} and Sang 2010) for the class $\mathcal{H}$. Let $P$ be a probability measure on $S$ and let $Z_{i}: S^{\mathbb{N}}\rightarrow S$ be the coordinate functions of $S^{\mathbb{N}}$, which are i.i.d. (P) and set $Pr=P^\mathbb{N}$.

As $\mathcal{H}$ is VC type, bounded and countable, then there exists 
$0<C_{j}<\infty$, $1\leq j\leq3$ depending on $ v(\mathcal{H})$ and $A_{1}$ such that, for all $t$ with 
\begin{align*}
C_{1}\sqrt{m}\sigma\sqrt{\log \frac{2\lVert K_{4} \rVert_{\infty}}{\sigma}}\leq t\leq \frac{m\sigma^2}{\lVert K_{4}\rVert_\infty},
\end{align*}
we have 
\begin{align}\label{TL}
Pr\left\{\max_{1\leq k\leq m}\left\lVert \sum_{j=1}^{k}(h_u(Z_{j})-\mathbb{E}h_u(Z_1))\right\rVert_{\mathcal{H}}>t\right\}\leq C_{2}\exp\left( -C_{3}\frac{t^2}{m\sigma^2}\right),
\end{align}
where
\begin{align*}
\lVert K_{4}\rVert_{\infty}>\sigma^2>\lVert Var(h_u)\rVert_{\mathcal{H}}.
\end{align*}
Here and in the rest of the paper we use the notations $\|f\|_\infty:=\sup|f(\cdot)|$ for some function $f$ and $\lvert\lvert\cdot\rvert\rvert_{\mathcal{H}}:=\sup_{h\in\mathcal{H}} \lvert\cdot\rvert$ for some class of functions $\mathcal{H}$. 
The proof of the following theorem has a similar pattern as the proof of a theorem in Gin\'{e} and Sang (2010) and it consists of blocking and application of Talagrand inequality. It extends to the optimal convergence rate and shows an ideal result after applying Borel-Cantelli lemma.

\begin{theorem}\label{T1}
Let $\mathcal{H}$ be the class of functions defined in equation  \eqref{H}. Assume that $\Vert h_u \Vert_{\infty}\leq K_{4}$ for some $0<K_{4}<\infty$. Then \begin{align}\label{TH}
\left\Vert \frac{1}{m}\sum_{j=1}^{m} h_{u}(Z_{j}) - \mathbb{E}h_{u}(Z_{1})\right\Vert_{\mathcal{H}}=O_{a.s}\left(\left(\frac{\log m}{m}\right)^\frac{1}{2}\right).
\end{align}

\end{theorem}
\begin{proof} We block the terms between dyadic integers as follows.
\begin{align*}
&Pr\left\{\max_{2^{k-1}<m\leq 2^k}\left( \frac{m}{\log m}  \right)^\frac{1}{2} \left\Vert \frac{1}{m}\sum_{j=1}^{m} h_{u}(Z_{j}) - \mathbb{E}h_{u}(Z_{1})\right\Vert_{\mathcal{H}}>\lambda\right\}\\
&=Pr\left\{\max_{2^{k-1}<m\leq 2^k}\sqrt{\frac{1}{m\log m}} \left\Vert \sum_{j=1}^{m} \left\{h_{u}(Z_{j}) - \mathbb{E}h_{u}(Z_{j})\right\}\right\Vert_{\mathcal{H}}>\lambda\right\}\\
&\le Pr\left\{\max_{2^{k-1}<m\leq 2^k}\sqrt{\frac{1}{2^{k-1}\log 2^{k-1}}} \left\Vert \sum_{j=1}^{m} \left\{h_{u}(Z_{j}) - \mathbb{E}h_{u}(Z_{j})\right\}\right\Vert_{\mathcal{H}}>\lambda\right\} \\
&\le Pr\left\{\max_{2^{k-1}<m\leq 2^k}\left\Vert \sum_{j=1}^{m} \left\{h_{u}(Z_{j}) - \mathbb{E}h_{u}(Z_{j})\right\}\right\Vert_{\mathcal{H}}>\lambda \sqrt{2^{k-1}(k-1)}  \right\} \text{for any }\lambda>0.
\end{align*}
As we see from Lemma \ref{lemma2} and lemma \ref{lemma3}  that $\mathcal{H}=\bigg\{h_{u}:h_{u}(z)= \phi(1-f_{w}(g_{\theta}(z))) \bigg\}$ is a bounded VC class of measurable functions with respect to the constant envelope $K_{4}$. Hence, the subclasses of $\mathcal{H}$ are VC classes of functions with respect to $K_{4}$ and with the same characteristics $A_{1}$ and $v(\mathcal{H})$.
Next, applying Talagrand inequality from \eqref{TL}, we obtain
\begin{align*}
&Pr\left\{\max_{2^{k-1}<m\leq 2^k}\left\Vert \sum_{j=1}^{m} h_{u}(Z_{j}) - \mathbb{E}h_{u}(Z_{j})\right\Vert_{\mathcal{H}}>\lambda \sqrt{2^{k-1}(k-1)}\right\}\leq C_{2} \exp\left( -C_{3}\frac{\lambda^2 2^{k-1}(k-1)}{m\sigma^2}\right) \\
&=Pr\left\{\max_{2^{k-1}<m\leq 2^k}\left\Vert \sum_{j=1}^{m} h_{u}(Z_{j})-\mathbb{E}h_{u}(Z_{j})\right\Vert_{\mathcal{H}}>\lambda \sqrt{2^{k-1}(k-1)}\right\}\leq C_{2} \exp\left( -C_{3}\frac{\lambda^2 2^{k-1}(k-1)}{2^k\sigma^2}\right).
\end{align*}
Let \begin{align*} A_{k}=\left\{\max_{2^{k-1}<m\leq 2^k}\left\Vert \frac{1}{m}\sum_{j=1}^{m} h_{u}(Z_{j}) - \mathbb{E}h_{u}(Z_{j})\right\Vert_{\mathcal{H}}>\lambda \sqrt{2^{k-1}(k-1)}\right\}.
\end{align*}
Then using the Borel-Cantelli lemma we get
\begin{align*}
&\sum_{k=1}^{\infty}Pr(A_{k})\leq C_{2} \sum_{k=1}^{\infty}\exp\left( -C_{3}\frac{\lambda^2 2^{k-1}(k-1)}{2^k\sigma^2}\right)\\
&\leq C_{2} \sum_{k=1}^{\infty}\exp\left( -C_{3}\frac{\lambda^2(k-1)}{2\sigma^2}\right) <\infty.
\end{align*}
Hence we can write 
\begin{align*}
Pr(\limsup A_{k})=0.
\end{align*}
\end{proof}

Similarly, $\mathcal{F}_{1}$ is VC type, bounded and countable. Then there exists 
$0<C_{i}<\infty$, $1\leq i\leq3$ depending on $ v(\mathcal{F}_{1})$ and $A_{2}$ such that, for all $t$ with 
\begin{align*}
C_{1}\sqrt{n}\sigma\sqrt{\log \frac{2\lVert K_{3} \rVert_{\infty}}{\sigma}}\leq t\leq \frac{n\sigma^2}{\lVert K_{3}\rVert_\infty},
\end{align*}
we have 
\begin{align}\label{TLF1}
Pr\left\{\max_{1\leq k\leq n}\left\lVert \sum_{i=1}^{k}(f_{1}(X_{i})-\mathbb{E}f_{1}(X_{i}))\right\rVert_{\mathcal{F}_{1}}>t\right\}\leq C_{2}\exp\left( -C_{3}\frac{t^2}{n\sigma^2}\right),
\end{align}
where
\begin{align*}
\lVert K_{3}\rVert_{\infty}>\sigma^2>\lVert Var(f_{1})\rVert_{\mathcal{F}_{1}}.
\end{align*}
\begin{theorem}\label{T2}
Let $\mathcal{F}_{1}$ be the class of functions defined in  \eqref{F1}. Assume that $\Vert f_{1} \Vert_{\infty}\leq K_{3}$, for some $0<K_{3}<\infty$. Then 
\begin{align}\label{TF1}
\left\Vert \frac{1}{n}\sum_{i=1}^{n} f_{1}(X_{i}) - \mathbb{E}f_{1}(X_{i})\right\Vert_{\mathcal{F}_1}=O_{a.s}\left(\left(\frac{\log n}{n}\right)^\frac{1}{2}\right).
\end{align}
\end{theorem}	
\noindent \textbf{The Proof of Theorem \ref{theorem1}}.

\noindent  Add equation \eqref{TH} and \eqref{TF1} from Theorem \ref{T1} and \ref{T2} then put in Lemma \ref{L2.1}, we have Theorem \ref{theorem1}.


\section{Generalization on the existing error bounds}\label{exten}
In this section, we study the existing errors using the technique applied in the previous section. The technique is applied for general objective function from Arora et al. (2017) and neural network distance. For the general objective function and neural network distance we get almost sure result. The error bound for neural network distance is the particular case of the objective function \eqref{sup1}.

\subsection{Error for general objective function}
Since for any function $f_w$ in $\mathcal{F}$, 
$\left\vert f_{w}(X_{1})-f_{w}(X_{2})\right\vert\le U_{w}\|X_{1}-X_{2}\|$ where $U_{w}=\prod_{i=1}^{d} M_{w}(i) \prod_{i=1}^{d-1} L_{w}(i)$ which can be derived as in the proof of Lemma \ref{lemma3}. Therefore, $\mathcal{F}=\{f_{w}:w\in\mathcal{W}\}$ is a class of $U_{w}$-Lipschitz functions.
If $g_\theta^{\hat{\mu}_m}$ is the empirical distribution of the generator and $\hat{\nu}_{n}$ is the empirical distribution  of  the target distribution $\nu$ then Arora et al. (2017) defined GAN objective function as \eqref{sup1},
where $\phi$ is the measuring function which takes values in $[-\Delta,\Delta]$ and is $L_{\phi}$-Lipschitz.
With i.i.d. observations $X_1, \cdots, X_n\sim \nu$ and generated data $g_{\theta}(Z)\sim g_\theta^{\mu}$,  the empirical version of \eqref{sup1} is given by \eqref{sup2}.\\

They derived error bound for
$ \left \vert d_{\mathcal{F,\phi}}(g_\theta^{\hat{\mu}_m},\hat\nu_{n})-d_{\mathcal{F,\phi}}(g_\theta^{{\mu}},\nu)\right\vert$. In their paper, they fixed a generator function and only considered the discriminator class while we work on  both discriminator and generator classes of functions which are indexed by bounded weight matrices and vectors.
In fact,  our technique can also be applied to bound the generalization error studied in Arora et al. (2017). We have the following corollary for a similar generalization error.
\begin{corollary}
Let $\mathcal{F}$ be the discriminator class of functions. Let $g_\theta^{\hat{\mu}_m}$ be the empirical distribution of the generator and $\hat{\nu}_{n}$ be the empirical distribution  of  the target distribution $\nu$. Under the assumptions as in Theorem \ref{theorem1}, we have 
\begin{align}\label{C4.2}
\sup_{\theta\in \Theta}\left \vert d_{\mathcal{F,\phi}}(g_\theta^{\hat{\mu}_m},\hat\nu_{n})-d_{\mathcal{F,\phi}}(g_\theta^{{\mu}},\nu)\right\vert=O_{a.s}\left(\left(\frac{\log m}{m}\right)^\frac{1}{2}\right)+O_{a.s}\left(\left(\frac{\log n}{n}\right)^\frac{1}{2}\right).
\end{align}
\end{corollary}
\begin{proof}
We can write the bound according to Lemma \ref{L2.1}.
\begin{align*}
&\sup_{\theta\in \Theta}\left \vert d_{\mathcal{F,\phi}}(g_\theta^{\hat{\mu}_m},\hat\nu_{n})-d_{\mathcal{F,\phi}}(g_\theta^{{\mu}},\nu)\right\vert\\
&=\sup_{\theta\in \Theta}\left\vert  \sup_{w\in\mathcal{W}}\left\vert\hat{\mathbb{E}}_n \phi(f_w(X))+\hat{\mathbb{E}}_m\phi (1-f_w(g_{\theta}(Z)))\right\vert-2\phi(1/2) \right.\\
&\left.- \sup_{w\in \mathcal{W}}\left\vert{\mathbb{E}} \phi(f_w(X))+{\mathbb{E}}\phi (1-f_w(g_{\theta}(Z)))\right\vert-2\phi(1/2)\right\vert \\
&\leq\sup_{\theta\in \Theta} \sup_{w\in \mathcal{W}}\left \vert \frac{1}{m}\sum_{j=1}^{m}\phi (1-f_{w}(g_{\theta}(Z_{j})))+\frac{1}{n}\sum_{i=1}^{n}\phi (f_{w}(X_{i}))- [{\mathbb{E}}\phi (1-f_w(g_{\theta}(Z)))+{\mathbb{E}} \phi(f_w(X))]\right\vert\\
&\leq \sup_{\theta\in \Theta} \sup_{w\in \mathcal{W}}\left \vert \frac{1}{m}\sum_{j=1}^{m}\phi(1-f_{w}(g_{\theta}(Z_{j})))- \mathbb{E}\phi (1-f_{w}(g_{\theta}(Z)))\right\vert+ \sup_{w\in \mathcal{W}}\left \vert \frac{1}{n}\sum_{i=1}^{n}\phi(f_{w}(X_{i}))-\mathbb{E}\phi(f_{w}(X))\right\vert.
\end{align*}
 For $u=(\theta,w)$,
 $h_u(Z)=\phi(1-f_{w}(g_{\theta}(Z)))$ and $f_{1}=\phi(f_{w}(X))$, we have 
\begin{align}\label{H2}
&\sup_{\theta\in \Theta}\left \vert d_{\mathcal{F,\phi}}(g_\theta^{\hat{\mu}_m},\hat\nu_{n})-d_{\mathcal{F,\phi}}(g_\theta^{{\mu}},\nu)\right\vert\nonumber\\
\leq &\sup_{u\in U} \left \vert\frac{1}{m}\sum_{j=1}^{m} \left( h_u(Z_{j})-\mathbb{E}h_u(Z_{j})\right)\right\vert +\sup_{w\in \mathcal{W}} \left \vert\frac{1}{n}\sum_{i=1}^{n} \left( f_{1}(X_{i})-\mathbb{E}f_{1}(X_{i})\right)\right\vert.  
\end{align}
	
The inequality \eqref{H2} gives same bound in Lemma \ref{L2.1}. 
Then from Theorem \ref{T1} and \ref{T2} we get the bound $O_{a.s}\left(\left(\frac{\log m}{m}\right)^\frac{1}{2}\right)+O_{a.s}\left(\left(\frac{\log n}{n}\right)^\frac{1}{2}\right)$.
\end{proof}

\begin{remark}
Arora et al. (2017) derived the following bound. With probability at least $1-e^{-p}$, for all $n$ with
$n\ge \frac{cp\Delta^{2}\log(U_{w}L_{\phi}p/\epsilon)}{\epsilon^2}$, here $c$ is a constant,
\begin{align}\label{R4}
\left \vert d_{\mathcal{F,\phi}}(g_\theta^{\hat{\mu}_m},\hat\nu_{n})-d_{\mathcal{F,\phi}}(g_\theta^{{\mu}},\nu)\right\vert\le \epsilon,
\end{align}
where $p$ denotes the number of parameters in $w$ and recall that $\Delta, U_{w}, L_{\phi}$ was introduced at the beginning of this section. If $\epsilon=\sqrt{\frac{cp\Delta^{2}\log(U_{w}L_{\phi}p/\epsilon)}{n}}$
our bound \eqref{C4.2} is consistent with \eqref{R4}. However, \eqref{R4} holds with probability $1-e^{-p}$ and with fixed generator class while in this paper we have the almost sure result and we incorporate both discriminator and generator with sample sizes $n$ and $m$ respectively.
\end{remark}

\subsection{Error for neural network distance}
When $\phi(x) = x$, the expression $d_{\mathcal{F,\phi}}(g_\theta^{{\mu}},\nu)$ from equation \eqref{sup1}, with a slight abuse of notation, can be written as following:
\begin{equation}\label{sup4}
d_{\mathcal{F}}(g_\theta^{{\mu}},\nu)=\sup _{w\in \mathcal{W}} [\mathbb{E}f_{w}(g_{\theta}(Z))-\mathbb{E}f_{w}(X)]. 
\end{equation}
This quantity is known as the neural network distance, as cited in Arora et al. (2017). To estimate the second expectation, one employs empirical average and minimize the following empirical version of \eqref{sup4}:
\begin{equation}\label{sup5}
d_{\mathcal{F}}(g_\theta^{{\mu}},\hat\nu_{n})= \sup_{w\in \mathcal{W}}\left[\mathbb{E}f_{w}(g_{\theta}(Z))-\frac{1}{n}\sum_{i=1}^{n}f_{w}(X_{i})\right].
\end{equation}
Zhang et al. (2018) discussed the generalization error of GAN model within the framework of equations \eqref{sup4} and \eqref{sup5}. They considered GAN training as:
\begin{align}\label{inf2}
\inf_{\theta\in\Theta} d_{\mathcal{F}}(g_\theta^{{\mu}},\hat\nu_{n}).
\end{align}

In particular, if $\hat\theta$ is the solution of the equation \eqref{inf2} that satisfies
$d_{\mathcal{F}}(g_{\hat\theta}^{{\mu}},\hat\nu_{n})\le\inf\limits_{\theta\in\Theta}d_{\mathcal{F}}(g_\theta^{{\mu}},\hat\nu_{n})+ \epsilon$, $\epsilon\ge0$, 
they developed bound for $ d_{\mathcal{F}}(g_{\hat\theta}^{{\mu}},\nu)-\inf\limits_{\theta\in\Theta}d_{\mathcal{F}}(g_\theta^{{\mu}},\nu) $. We are working on a different framework of error bound 
$\inf\limits_{\theta\in\Theta}d_{\mathcal{F,\phi}}(g_\theta^{{\hat\mu_m}},\hat\nu_{n})-\inf\limits_{\theta\in\Theta}d_{\mathcal{F,\phi}}(g_\theta^{{\mu}},\nu),$
which is given by  the difference of the empirical version and the population version. However, we can also apply our technique to the framework in Zhang et al. (2018) and find error bound which is given by  Corollary \ref{C1}. On the other hand, we obtain almost sure convergence rate which depends on sample sizes of both discriminator and generator where their bound depends on Rademacher complexity and discriminator sample size.
\begin{corollary}\label{C1}
Let $\mathcal{F}$ be the discriminator class of functions. Assume that $\phi$ is monotone increasing and $X$ has bounded norm, $\|X\|<\infty$. Let $\hat\theta$ be the solution of\;  $\inf\limits_{\theta\in\Theta}d_{\mathcal{F},\phi}(g_\theta^{{\mu}},\hat\nu_{n})$ that satisfies $d_{\mathcal{F},\phi}(g_{\hat\theta}^{{\mu}},\hat\nu_{n})\le\inf\limits_{\theta\in\Theta}d_{\mathcal{F},\phi}(g_\theta^{{\mu}},\hat\nu_{n})+ \epsilon$.
Then 
\begin{align}\label{C2}
d_{\mathcal{F},\phi}(g_{\hat\theta}^{{\mu}}, \nu)-\inf_{\theta\in\Theta}d_{\mathcal{F},\phi}(g_\theta^{{\mu}}, \nu)=O_{a.s}\left(\left(\frac{\log n}{n}\right)^\frac{1}{2}\right)+\epsilon,
\end{align}
where $d_{\mathcal{F},\phi}(g_\theta^{{\mu}},\nu)$ and $d_{\mathcal{F},\phi}(g_\theta^{{\mu}},\hat\nu_{n})$ are given in \eqref{sup1} and \eqref{sup3}.
\end{corollary}
\begin{proof}
We can find the bound as follows	
\begin{align*}
&d_{\mathcal{F},\phi}(g_{\hat\theta}^{{\mu}}, \nu)-\inf_{\theta\in\Theta}d_{\mathcal{F},\phi}(g_\theta^{{\mu}}, \nu)\\
&=d_{\mathcal{F},\phi}(g_{\hat\theta}^{{\mu}}, \nu)-d_{\mathcal{F},\phi}(g_{\hat\theta}^{{\mu}},\hat\nu_{n})+d_{\mathcal{F},\phi}(g_{\hat\theta}^{{\mu}},\hat\nu_{n})- \inf_{\theta\in\Theta}d_{\mathcal{F},\phi}(g_\theta^{{\mu}}, \nu)\\
&\le d_{\mathcal{F},\phi}(g_{\hat\theta}^{{\mu}}, \nu)-d_{\mathcal{F},\phi}(g_{\hat\theta}^{{\mu}},\hat\nu_{n})+\inf_{\theta\in\Theta}d_{\mathcal{F},\phi}(g_{\hat\theta}^{{\mu}},\hat\nu_{n})- \inf_{\theta\in\Theta}d_{\mathcal{F},\phi}(g_\theta^{{\mu}}, \nu)+\epsilon\\
&\le 2 \sup_{\theta\in\Theta}\left \vert d_{\mathcal{F},\phi}(g_\theta^{{\mu}},\nu)-d_{\mathcal{F},\phi}( g_\theta^{{\mu}},\hat \nu_{n})\right\vert+\epsilon\\
&= 2\sup_{\theta\in\Theta}\left\vert \sup _{w\in\mathcal{W}} \left\vert\mathbb{E}\phi(f_{w}(X))+\mathbb{E}\phi (1-f_{w}(g_{\theta}(Z)))\right\vert-2\phi(1/2) \right.\\
&\left.- \sup_{w\in \mathcal{W}}\left\vert\hat{\mathbb{E}}_n \phi(f_w(X))+{\mathbb{E}}\phi (1-f_w(g_{\theta}(Z)))\right\vert-2\phi(1/2)\right\vert+\epsilon \\
&\leq  2\sup _{w\in \mathcal{W}}\lvert\mathbb{E}\phi(f_{w}(X))-\hat{\mathbb{E}}_{n}\phi(f_{w}(X))\rvert+\epsilon\\
&=  2\sup_{w\in \mathcal{W}} \left\lvert \left[\frac{1}{n}\sum_{i=1}^{n} \phi(f_{w}(X_{i}))- \mathbb{E}\phi(f_{w}(X))\right]\right\rvert+\epsilon\\
&=  2\sup_{w\in\mathcal{W}} \left\vert\frac{1}{n}\sum_{i=1}^{n} \left(f_{1}(X_{i})-\mathbb{E}f_{1}(X_{i})\right)\right\vert+\epsilon\;. 
\end{align*}
The structure of the discriminator class $\mathcal{F}:=\{ f_{w}(x):w\in\mathcal{W} \}$  is taken from \eqref{F} with weight parameters defined in \eqref{W}. The corresponding VC-dimension and bound of $\mathcal{F}_{1}:=\{f_{1}=\phi(f_{w}(x)):w\in \mathcal{W} \}$ are given in \eqref{VCF1} and \eqref{ENFP} respectively.
\begin{align*}
v(\mathcal{F}_{1})=O\left(d\sum_{i=1}^{d}\lVert W_{i}\rVert_{0}\log\left(\sum_{i=1}^{d}\lVert W_{i}\rVert_{0}\right)\right),
\end{align*}
\begin{align*}
\vert \mathcal{F}_{1}\vert\le K_{3}.
\end{align*}
Therefore, we have the Talagrand inequality for  $\mathcal{F}_{1}:=\{\phi(f_{w}(x)):w\in \mathcal{W} \}$ as \eqref{TLF1}. Then using  Theorem \ref{T2} the result \eqref{C2} is proved.
\end{proof}

\begin{remark}
With probability at least $1-\delta$, Zhang et al. (2018) derived the following bound  
\begin{align}\label{R1}
d_{\mathcal{F}}(g_{\hat\theta}^{{\mu}}, \nu)-\inf_{\theta\in\Theta}d_{\mathcal{F}}(g_\theta^{{\mu}}, \nu)\le2 \mathcal{R}_{n}(\mathcal{F})+2K_{1}\sqrt{\frac{2\log\left(1/\delta\right)}{n}}+\epsilon,
\end{align}
where the Rademacher complexity of $\mathcal{F}$ is defined as
\begin{align}\label{Red}
\mathcal{R}_{n}(\mathcal{F})=\mathbb{E}\left[  \sup_{w\in \mathcal{W}}\frac{2}{n}\sum_{i}\tau_{i}f_{w}(X_{i})\right].
\end{align}
Here the Rademacher variable $\tau_{i}$ is defined as $P(\tau_{i}=1)=P(\tau_{i}=-1)=\frac{1}{2}$ and $\lvert f_{w}\rvert\le K_{1}$. If we compare \eqref{R1} and \eqref{C2},  \eqref{R1} has  an additional term $\mathcal{R}_{n}(\mathcal{F})$.  Besides, they have the upper bound \eqref{R1} with the probability $1-\delta$ where our bound holds almost surely. Moreover, \eqref{C2} is for the general objective function, while the error defined in \eqref{R1} is the particular case with $\phi(x)=x$.
\end{remark}

Ji et al. (2021) considered the same case as ours that incorporates both of the  discriminator and generator neural networks with corresponding weights and input variables bounded. They defined GANs training as 
\begin{align*}
\inf_{\theta\in\Theta} d_{\mathcal{F}}(g_\theta^{{\hat\mu_m}},\hat\nu_{n}), 
\end{align*}
where $d_{\mathcal{F}}(g_\theta^{{\hat\mu_m}},\hat\nu_{n})$ can be evaluated from \eqref{sup2} with a slight change of the notation in the case of $\phi(x)=x$ as
\begin{equation*}
d_{\mathcal{F}}(g_\theta^{{\hat\mu_m}},\hat\nu_{n})=\sup_{w\in \mathcal{W}}[\hat{\mathbb{E}}_m f_w(g_{\theta}(Z))-\hat{\mathbb{E}}_n f_w(X)],
\end{equation*}
i.e., 
\begin{equation*}
d_{\mathcal{F}}(g_\theta^{{\hat\mu_m}},\hat\nu_{n})= \sup_{w\in\mathcal{W}}\left[\frac{1}{m}\sum_{j=1}^{m}f_{w}(g_{\theta}(Z_{j}))-\frac{1}{n}\sum_{i=1}^{n}f_{w}(X_{i})\right]. 
\end{equation*}
Let $\theta_{1}=\arg\inf\limits_{\theta\in\Theta} d_{\mathcal{F}}(g_\theta^{{\hat\mu_m}},\hat\nu_{n})$, then they used $ d_{\mathcal{F}}(g_{\theta_{1}}^{{\mu}},\nu)-\inf\limits_{\theta\in \Theta}d_{\mathcal{F}}(g_\theta^{{\mu}},\nu)$ as the definition of estimation error where $d_{\mathcal{F}}(g_\theta^{{\mu}},\nu)$ is given in \eqref{sup4}. However, their error bound involves Rademacher complexity while the bound in this paper does not involve it. Besides, we give almost surely convergence rates on the bound. The following corollary shows that the error defined in Ji et al. (2021) has almost sure bound by the technique in this paper for any monotone increasing function $\phi(x)$.
 
\begin{corollary}\label{Ji1}
 Under the setup and assumptions as in Theorem \ref{theorem1}, 
 let $\theta_{1}$ be the solution of $\inf\limits_{\theta\in\Theta} d_{\mathcal{F,\phi}}(g_\theta^{{\hat\mu_m}},\hat\nu_{n})$ where $d_{\mathcal{F,\phi}}(g_\theta^{{\hat\mu_m}},\hat\nu_{n})$ is given in \eqref{sup2}, we have 
\begin{align}\label{Ji2}
 d_{\mathcal{F,\phi}}(g_{\theta_{1}}^{{\mu}},\nu)-\inf\limits_{\theta\in \Theta}d_{\mathcal{F,\phi}}(g_\theta^{{\mu}},\nu)=O_{a.s}\left(\left(\frac{\log m}{m}\right)^\frac{1}{2}\right)+O_{a.s}\left(\left(\frac{\log n}{n}\right)^\frac{1}{2}\right).
\end{align}
\end{corollary}
\begin{proof} We get a decomposition of $ d_{\mathcal{F,\phi}}(g_{\theta_{1}}^{{\mu}},\nu)-\inf\limits_{\theta\in \Theta}d_{\mathcal{F,\phi}}(g_\theta^{{\mu}},\nu)$ from the proof of Theorem 1 in Ji et al. (2021) which is given as,
\begin{align}
&d_{\mathcal{F,\phi}}(g_{\theta_{1}}^{{\mu}},\nu)-\inf\limits_{\theta\in \Theta}d_{\mathcal{F,\phi}}(g_\theta^{{\mu}},\nu) \notag\\
&=d_{\mathcal{F,\phi}}(g_{\theta_{1}}^{{\mu}},\nu)-d_{\mathcal{F},\phi}(g_{\theta_{1}}^{{\mu}},\hat\nu_{n})+\inf_{\theta\in\Theta}d_{\mathcal{F},\phi}(g_{\theta}^{{\mu}},\hat\nu_{n})-\inf_{\theta\in\Theta}d_{\mathcal{F},\phi}(g_{\theta}^{{\mu}},\nu)+d_{\mathcal{F},\phi}(g_{\theta_{1}}^{{\mu}},\hat\nu_{n})-\inf_{\theta\in\Theta}d_{\mathcal{F},\phi}(g_{\theta}^{{\mu}},\hat\nu_{n}) \label{123th}.
\end{align}
The first part can be written as 
\begin{align}
& d_{\mathcal{F},\phi}(g_{\theta_{1}}^{{\mu}},\nu)-d_{\mathcal{F},\phi}(g_{\theta_{1}}^{{\mu}},\hat\nu_{n}) \notag\\
&=\sup _{w\in \mathcal{W} } \left\vert\mathbb{E}\phi (1-f_{w}(g_{\theta_{1}}(Z)))+\mathbb{E}\phi(f_{w}(X))\right\vert-\sup_{w\in\mathcal{W} }\left\vert\mathbb{E}\phi (1-f_{w}(g_{\theta_{1}}(Z)))+\frac{1}{n}\sum_{i=1}^{n}\phi (f_{w}(X_{i}))\right\vert \label {}\notag\\
&\le\sup _{w\in\mathcal{W} } \left\vert\mathbb{E}\phi(f_{w}(X))-\frac{1}{n}\sum_{i=1}^{n}\phi (f_{w}(X_{i}))\right\vert. \label{1st}
\end{align}
Let $\theta_{2}=\arg\min\limits_{w\in\mathcal{W}}d_{\mathcal{F},\phi}(g_{\theta}^{{\mu}},\nu)$. Then we obtain for the second part, 
\begin{align}
\inf_{\theta\in\Theta}d_{\mathcal{F},\phi}(g_{\theta}^{{\mu}},\hat\nu_{n})-\inf_{\theta\in\Theta}d_{\mathcal{F},\phi}(g_{\theta}^{{\mu}},\nu)\le
d_{\mathcal{F},\phi}(g_{\theta_{2}}^{{\mu}},\hat\nu_{n})-d_{\mathcal{F},\phi}(g_{\theta_{2}}^{{\mu}},\nu) \le \sup _{w\in\mathcal{W} } \left\vert\mathbb{E}\phi(f_{w}(X))-\frac{1}{n}\sum_{i=1}^{n}\phi (f_{w}(X_{i}))\right\vert. \label{2nd}
\end{align}
If $\hat\theta=\arg\min\limits_{w\in\mathcal{W}}d_{\mathcal{F},\phi}(g_{\theta}^{{\mu}},\hat\nu_{n})$,
we obtain for the third part as,
\begin{align}
& d_{\mathcal{F},\phi}(g_{\theta_{1}}^{{\mu}},\hat\nu_{n})-\inf_{\theta\in\Theta}d_{\mathcal{F},\phi}(g_{\theta}^{{\mu}},\hat\nu_{n}) \notag\\
&\le d_{\mathcal{F},\phi}(g_{\theta_{1}}^{{\mu}},\hat\nu_{n})-d_{\mathcal{F},\phi}(g_{\theta_{1}}^{{\hat\mu_{m}}},\hat\nu_{n})+d_{\mathcal{F},\phi}(g_{\hat\theta}^{{\hat\mu_{m}}},\hat\nu_{n})-d_{\mathcal{F},\phi}(g_{\hat\theta}^{{\mu}},\hat\nu_{n}) \notag\\
&\le \sup_{w\in \mathcal{W}}\left \vert \frac{1}{m}\sum_{j=1}^{m}\phi(1-f_{w}(g_{\theta_{1}}(Z_{j})))- \mathbb{E}\phi (1-f_{w}(g_{\theta_{1}}(Z)))\right\vert  \notag \\ 
&+ \sup_{w\in \mathcal{W}}\left \vert \frac{1}{m}\sum_{j=1}^{m}\phi(1-f_{w}(g_{\hat\theta}(Z_{j})))- \mathbb{E}\phi (1-f_{w}(g_{\hat\theta}(Z)))\right\vert \notag\\
&\le 2\sup_{\theta\in \Theta} \sup_{w\in \mathcal{W}}\left \vert \frac{1}{m}\sum_{j=1}^{m}\phi(1-f_{w}(g_{\theta}(Z_{j})))- \mathbb{E}\phi (1-f_{w}(g_{\theta}(Z)))\right\vert \label{3rd} .  
\end{align}
Using \eqref{1st}, \eqref{2nd} and \eqref{3rd} in \eqref{123th} we obtain, 
\begin{align*}
&d_{\mathcal{F},\phi}(g_{\theta_{1}}^{{\mu}},\nu)-\inf\limits_{\theta\in \Theta}d_{\mathcal{F},\phi}(g_\theta^{{\mu}},\nu)\\ 
&\le 2\sup_{\theta\in \Theta} \sup_{w\in \mathcal{W}}\left \vert \frac{1}{m}\sum_{j=1}^{m}\phi(1-f_{w}(g_{\theta}(Z_{j})))- \mathbb{E}\phi (1-f_{w}(g_{\theta}(Z)))\right\vert+2\sup_{w\in \mathcal{W}}\left \vert \frac{1}{n}\sum_{i=1}^{n}\phi(f_{w}(X_{i}))-\mathbb{E}\phi(f_{w}(X))\right\vert\\
&= 2\sup_{u\in U} \left \vert\frac{1}{m}\sum_{j=1}^{m} \left( h_{u}(Z_{j})-\mathbb{E}h_{u}(Z_{j})\right)\right\vert 
+2\sup_{w\in \mathcal{W}} \left \vert\frac{1}{n}\sum_{i=1}^{n} \left( f_{1}(X_{i})-\mathbb{E}f_{1}(X_{i})\right)\right\vert
\end{align*}
which is the same bound as in Lemma \ref{L2.1}. So we can find corresponding VC-dimension, envelop function and applying the Talagrand inequality and obtain the bound $O_{a.s}\left(\left(\frac{\log m}{m}\right)^\frac{1}{2}\right)+O_{a.s}\left(\left(\frac{\log n}{n}\right)^\frac{1}{2}\right)$
as in Theorem \ref{theorem1}.
\end{proof} 

\begin{remark}
The upper bound in Ji et al. (2021) can be written as below. With probability at least $1-2\delta$, 
\begin{align}\label{Ji3}
 d_{\mathcal{F}}(g_{\theta_{1}}^{{\mu}}, \nu)-\inf_{\theta\in \Theta}d_{\mathcal{F}}(g_{\theta}^{{\mu}},\nu)
\le4 \mathcal{R}_{n}(\mathcal{F})+4\mathcal{R}_{n}(\mathcal{F}\times\mathcal{G})+2U_{w}\sqrt{2\log\frac{1}{\delta}}\left( \frac{B_X}{\sqrt{n}}+\frac{B_{Z}U_{v}}{\sqrt{m}}\right),
\end{align}
where $U_{w}=\prod_{i=1}^{d}M_{w}(i)\prod_{i=1}^{d-1} L_{w}(i)$
and $U_{v}=\prod_{j=1}^{s}M_{\theta}(j) \prod_{j=1}^{s-1} L_{\theta}(j)$ from \eqref{ENF} and \eqref{ENFG},  $\mathcal{R}_{n}(\mathcal{F})$ is given in \eqref{Red}
and for the composition function class $\mathcal{F}\circ\mathcal{G}$, the Rademacher complexity is
\begin{align*}
\mathcal{R}_{n}(\mathcal{F}\times\mathcal{G} )=\mathbb{E}\left[  \sup_{w\in\mathcal{W},\theta\in \Theta}\frac{2}{n}\sum_{j}\tau_{j}f_{w}(g_{\theta}(Z_{j}))\right].
\end{align*}
The bound in \eqref{Ji3} has additional terms $\mathcal{R}_{n}(\mathcal{F})$ and $\mathcal{R}_{n}(\mathcal{F}\times\mathcal{G})$. Furthermore, \eqref{Ji3} depends on certain probability $1-2\delta$ while \eqref{Ji2} gives the almost sure convergence rates. Moreover, $ d_{\mathcal{F}}(g_{\theta_{1}}^{{\mu}}, \nu)-\inf\limits_{\theta\in \Theta}d_{\mathcal{F}}(g_{\theta}^{{\mu}},\nu)$ is a special case of the error $ d_{\mathcal{F,\phi}}(g_{\theta_{1}}^{{\mu}}, \nu)-\inf\limits_{\theta\in \Theta}d_{\mathcal{F,\phi}}(g_{\theta}^{{\mu}},\nu)$ when $\phi(x)=x$. 
\end{remark}
Liang (2017) studied same type of generalization error for the objective function $d_{\mathcal{F}}(g_{\theta}^{{\mu}}, \nu)$ as in Zhang et al. (2018). He fixed generator function and only considered the discriminator class when the density function of the observations is in the Sobolev space. In his paper, error bound was derived for the GAN estimator $\hat \theta$= $\arg\inf\limits_{\theta\in\Theta}d_{\mathcal{F}}(g_{\theta}^{{\mu}}, \hat\nu_{n})$ where $d_{\mathcal{F}}(g_{\theta}^{{\mu}}, \hat\nu_{n})$ is given in \eqref{sup5}.
\begin{corollary}
Let $\mathcal{F}$ be the discriminator class of functions. Assume that $\phi$ is monotone increasing and $X$ has bounded norm, $\|X\|<\infty$. Let $\hat\theta$ be the solution of   $\inf\limits_{\theta\in\Theta}d_{\mathcal{F},\phi}(g_{\theta}^{{\mu}},\hat\nu_{n})$ where $d_{\mathcal{F,\phi}}(g_{\theta}^{{\mu}}, \hat\nu_{n})$ is given in \eqref{sup3}.
Then 
\begin{align}\label{C4}
\mathbb{E}d_{\mathcal{F},\phi}(g_{\hat\theta}^{{\mu}},\nu)-\inf_{\theta\in\Theta}d_{\mathcal{F},\phi}(g_{\theta}^{{\mu}},\nu)=O\left(\left(\frac{\log n}{n}\right)^\frac{1}{2}\right).
\end{align}
\end{corollary}
\begin{proof}
Let $\theta_{2}=\arg\min d_{\mathcal{F},\phi}(g_{\theta}^{{\mu}},\nu)$. With the same argument as \eqref{2nd}, the error bound can be simplified as following,
\begin{align*}
&\mathbb{E}d_{\mathcal{F},\phi}(g_{\hat\theta}^{{\mu}},\nu)-\inf_{\theta\in\Theta}d_{\mathcal{F},\phi}(g_{\theta}^{{\mu}},\nu)\\
&=\mathbb{E}\left[d_{\mathcal{F},\phi}(g_{\hat\theta}^{{\mu}},\nu)-\inf_{\theta\in\Theta}d_{\mathcal{F},\phi}(g_{\theta}^{{\mu}},\nu)\right]\\
&= \mathbb{E}\left[d_{\mathcal{F},\phi}(g_{\hat\theta}^{{\mu}},\nu)-d_{\mathcal{F},\phi}(g_{\hat\theta}^{{\mu}},\hat\nu_{n})+d_{\mathcal{F},\phi}(g_{\hat\theta}^{{\mu}},\hat\nu_{n})- \inf_{\theta\in\Theta}d_{\mathcal{F},\phi}(g_{\theta}^{{\mu}}, \nu)\right]\\
&\le \mathbb{E}\left[ d_{\mathcal{F},\phi}(g_{\hat\theta}^{{\mu}},\nu)-d_{\mathcal{F},\phi}(g_{\hat\theta}^{{\mu}},\hat\nu_{n})+d_{\mathcal{F},\phi}(g_{\theta_{2}}^{{\mu}},\hat\nu_{n})-d_{\mathcal{F},\phi}(g_{\theta_{2}}^{{\mu}}, \nu)\right]\\
&\le 2\mathbb{E} \sup_{w\in \mathcal{W}}\left\vert\frac{1}{n}\sum_{i=1}^{n}\phi(f_{w}(X_{i}))-{\mathbb{E}}\phi(f_w(X))\right\vert\\
&=2\mathbb{E} \sup_{w\in \mathcal{W}}\left\vert\frac{1}{n}\sum_{i=1}^{n}(f_{1}(X_{i})-{\mathbb{E}}f_{1}(X_{i}))\right\vert.
\end{align*}
Here, $\sup\limits_{w\in \mathcal{W}}\left\vert\frac{1}{n}\sum_{i=1}^{n}(f_{1}(X_{i})-{\mathbb{E}}f_{1}(X_{i}))\right\vert=O_{a.s}\left(\left(\frac{\log n}{n}\right)^\frac{1}{2}\right)$ by the same argument as in Theorem \ref{T2}. Hence we obtain \eqref{C4}.
\end{proof}
\begin{remark}
Liang (2017) derived bound as,
\begin{align} \label{Sob}
\mathbb{E}d_{\mathcal{F}}(g_{\hat\theta}^{{\mu}},\nu)-\inf_{\theta\in\Theta}d_{\mathcal{F}}(g_{\theta}^{{\mu}},\nu)\leq CL_{1}(k+1)^\frac{1}{2}(M_{w}(i))^d n^{-\frac{\alpha+1}{2\alpha+2+k}}, 
\end{align}
with the number of layer $d$, input $x\in [0,1]^k=\Omega$, weight matrix bounded by $\lVert W_{i}\rVert\leq M_{w}(i)$ and $d\nu_{x}/dx\in W^{\alpha,2}(L_{1})$ which is a Sobolev space with smoothness parameter $\alpha$. The Sobolev space is defined as,
\begin{align*}
W^{\alpha,2}(L_{1})=\left\{f_{w}: \Omega\rightarrow\mathbb{R},\left\vert\left\vert \mathcal{F}_{D}^{-1}\left[ (1+\lvert\vert\zeta\rvert\rvert^2)^{\alpha/2}\mathcal{F}_{D}f_{w}(\zeta)\right]\right\vert\right\vert_{2}\le L_{1} \right\}, 
\end{align*}
where $\mathcal{F}_D$ denotes Fourier transform and  $\mathcal{F}_D^{-1}$ is its inverse with $\zeta=(\zeta_{1},\zeta_{1}....\zeta_{k})$ the coefficients of Fourier series.
When comparing these two generalization error bounds in \eqref{C4} and \eqref{Sob}, we see that the factor $(M_{w}(i))^d$ appears in \eqref{Sob}.  Since $\frac{\alpha+1}{2\alpha+2+k}<\frac{1}{2}$ then   $\left(\frac{\log n}{n}\right)^\frac{1}{2}< n^{-\frac{\alpha+1}{2\alpha+2+k}}$ uniformly. Also \eqref{Sob} is the error defined for the specific case $\phi(x)=x$ in $\eqref{sup1}$.
\end{remark}

\section{Conclusion}\label{conc}
In this study, we have applied a novel error analysis technique that provides a tight upper bound for GANs by employing the Talagrand inequality. To establish the convergence rates of the bound, we also utilized the Borel-Cantelli lemma. Our investigation focuses on understanding the error properties of GANs with parameterized discriminator and generator and a general empirical objective function. The error rate is given in the almost sure sense. In addition, our technique has broad applications and can be used to bound existing errors. Some errors defined in the literature correspond to specific cases of the error defined in this paper.

We aim to explore various directions in the future endeavors. Primarily, computing the general objective function poses a challenging non-concave maximization problem. Hence, we are interested in delving into the theoretical analysis of these algorithms. Furthermore, we are keen on developing similar error analysis techniques for regularized GAN estimators, which can be an exciting avenue of research. On the other hand, we assume that the observations have bounded norm in this research. It is interesting to know whether one can remove this restriction in developing the upper bound of the error. Another challenging problem is to find the lower bound of the error and therefore one may claim the optimality of the error rate.

\noindent\textbf{Acknowledgement.} 
The research of Hailin Sang is partially supported by the Simons
Foundation Grant 586789, USA. 
\\

 \end{document}